\documentclass[10pt,twocolumn,letterpaper]{article}

\newcommand{\ours}{\textsc{Tangle}}

\newcommand{\genes}{\mathcal{G}}
\newcommand{\histology}{\mathcal{H}}
\newcommand{\real}{\mathbb{R}}

\usepackage{cvpr}              

\usepackage[accsupp]{axessibility}
\usepackage{url}
\usepackage{cite}
\usepackage{amsmath,amssymb,amsfonts}
\usepackage{algorithmic}
\usepackage{graphicx}
\usepackage{textcomp}
\usepackage{paralist}
\usepackage{listings}
\usepackage{booktabs}
\usepackage{hhline}
\usepackage{framed,multirow}
\usepackage{lipsum}  
\usepackage{soul}
\usepackage{times}
\usepackage{cancel}
\usepackage{booktabs}
\usepackage{latexsym}
\usepackage{array}
\usepackage{stackengine}
\usepackage{relsize}
\usepackage[ruled,vlined]{algorithm2e}

\makeatletter
\renewcommand*{\@fnsymbol}[1]{%
  \ensuremath{%
    \ifcase#1\or *
    \or \dagger
    \or \ddagger 
    \or \mathsection
    \or \mathparagraph
    \or \|\or **\or \dagger\dagger
    \or \ddagger\ddagger \else\@ctrerr\fi}%
}
\makeatother

\usepackage[dvipsnames]{xcolor}

\definecolor{cvprblue}{rgb}{0.21,0.49,0.74}
\usepackage[pagebackref,breaklinks,colorlinks,citecolor=cvprblue]{hyperref}

\def\paperID{7371} 
\def\confName{CVPR}
\def\confYear{2024}

\title{Transcriptomics-guided Slide Representation Learning\\ in Computational Pathology}

\author{Guillaume Jaume$^{1,2}\,$\thanks{Equal contribution}\;, Lukas Oldenburg$^{1,3\:*}$, Anurag Vaidya$^{1,2}$, Richard J. Chen$^{1,2}$, \\Drew F.K. Williamson$^{1,2\,}$\thanks{Presently at Emory University School of Medicine}\;, 
Thomas Peeters$^{1}$, Andrew H. Song$^{1,2}$, Faisal Mahmood$^{1,2}$\\
${^1}$Mass General Brigham, ${^2}$Harvard University and ${^3}$RWTH Aachen University\\
{\tt\small gjaume@bwh.harvard.edu, lukas.oldenburg@rwth-aachen.de, faisalmahmood@bwh.harvard.edu}
}

\begin{document}
\maketitle
\begin{abstract}
Self-supervised learning (SSL) has been successful in building \emph{patch embeddings} of small histology images (e.g., 224 $\times$ 224 pixels), but scaling these models to learn \emph{slide embeddings} from the entirety of giga-pixel whole-slide images (WSIs) remains challenging. Here, we leverage complementary information from gene expression profiles to guide slide representation learning using multimodal pre-training. Expression profiles constitute highly detailed molecular descriptions of a tissue that we hypothesize offer a strong task-agnostic training signal for learning slide embeddings. Our slide and expression (S+E) pre-training strategy, called $\ours$, employs modality-specific encoders, the outputs of which are aligned via contrastive learning. $\ours$ was pre-trained on samples from three different organs: liver (n=6,597 S+E pairs), breast (n=1,020), and lung (n=1,012) from two different species (\textit{Homo sapiens} and \textit{Rattus norvegicus}). Across three independent test datasets consisting of 1,265 breast WSIs, 1,946 lung WSIs, and 4,584 liver WSIs, $\ours$ shows significantly better few-shot performance compared to supervised and SSL baselines. When assessed using prototype-based classification and slide retrieval, $\ours$ also shows a substantial performance improvement over all baselines. Code available at \href{https://github.com/mahmoodlab/TANGLE}{https://github.com/mahmoodlab/TANGLE}. 
\end{abstract}

\section{Introduction}
\label{sec:intro}
 
Self-supervised learning (SSL)~\cite{caron2021emerging,zhou2022image} has recently gained significant traction in Computational Pathology (CPath)~\cite{song2023artificial,ciga2022self,shmatko2022artificial,wang2022transformer,chen2022scaling,mukashyaka2024ebiomedicine}. SSL is particularly suited for modeling giga-pixel whole-slide images (WSIs), whose size can exceed 150,000$\times$150,000 pixels, and which are consequently challenging to process with Vision Transformers (ViTs) or Convolutional Neural Networks (CNNs). Because of this size constraint, most CPath approaches adopt a divide-and-conquer strategy that consists of (1) tessellating the WSI into small patches and (2) extracting low-dimensional \emph{patch embeddings} with a frozen pre-trained network. Until recently, the prevalent practice involved relying on networks pre-trained on ImageNet~\cite{deng2009imagenet,he2016deep,lu2021data}. However, with the advent of SSL, this step is replaced by histopathology-specific visual encoders~\cite{wang2022transformer,filiot2023scaling,azizi2023robust,vorontsov2023virchow} or vision-language encoders~\cite{lu2023visual,huang2023visual}, in most cases trained on human cancer samples. The resulting patch embeddings constituting the WSI can then be fed to weakly-supervised models for classification as done in Multiple Instance Learning~\cite{dietterich1997solving,ilse2018attention,lu2021data,shao2021transmil,lee2022derivation}.

\begin{figure}[t]
   \includegraphics[width=\linewidth]{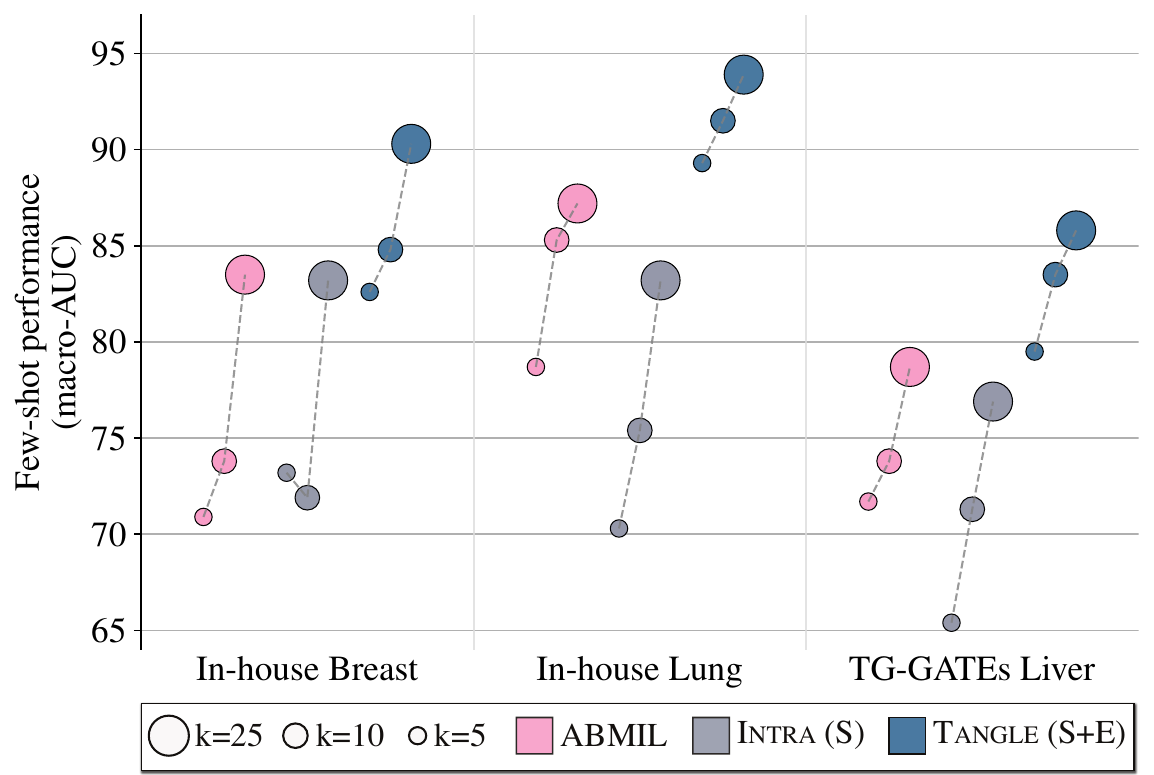}
   \caption{\textbf{Few-shot performance.} $\ours$ linear probing performance compared to multiple instance learning (ABMIL) and intra-modality slide SSL ($\textsc{Intra}$). $\ours$ uses gene expression (E) to guide slide pre-training (S) using multimodal contrastive learning (S+E). Results on independent cohorts for BRCA subtyping (human breast, n=1,265 WSIs), NSCLC subtyping (human lung, n=1,946 WSIs), and TG-GATEs lesion classification (rat liver, n=4,584 WSIs). k: number of training samples per class. 
   }
   \label{fig:title_figure_fewshot}
\end{figure}

SSL can also be pushed one step further to derive \emph{slide embeddings} without requiring any human annotations~\cite{koohbanani2021self,chen2022scaling,tavolara2022contrastive,lazard2023giga,yu2023slpd}. The resulting slide embeddings can serve as input for various downstream tasks with minimal or no training, enabling slide classification with few-shot learning and prototyping, slide retrieval, and case stratification. In addition, as the embedding space is learned without necessitating pathologist annotations, the risk of using noisy labels inherent in inter-observer variability is greatly mitigated~\cite{gomes2014inter}. However, building slide embeddings with SSL remains challenging as
(1) constructing slide ``views'' based on \emph{patch-level augmentations} requires extracting multiple patch embeddings per patch, which is computationally expensive;
(2) the visual primitives and invariances that need to be learned (such as being able to detect edges in natural images) become unclear when scaling to very large inputs; and
(3) intra-slide heterogeneity can prevent deriving a consistent and strong training signal, especially when using masked image modeling.

Instead, inspired by multimodal vision-language models, we leverage gene expression data to \emph{guide} slide representation learning into a slide-expression (S+E) pre-training model. Gene expression data, such as measured with RNA sequencing, are known to be strong indicators of disease state, with molecular signatures predictive of cancer subtype~\cite{mostavi2019convolutional}, patient survival~\cite{beer2002gene}, and drug toxicity~\cite{dann2016developments}, among others. Intuitively, the histology slide (S) and corresponding expression data (E) provide different \emph{views} of the same underlying biological processes: gene expression forms a highly detailed molecular description of tissue, with as many descriptors as there are transcriptomic measurements, albeit lacking spatial information. Conversely, histology slides offer a finely detailed spatial representation of the tissue but with only two markers, namely, the hematoxylin and eosin combination represented as RGB channels. Consequently, molecular alterations, as detected through bulk transcriptomics, can be exhibited as discernible morphological patterns when examining the associated histology slides~\cite{coudray2018classification,kather2019deep,kather2020pan}. We hypothesize that guiding slide representation learning with expression constitutes a much stronger training signal than using slide augmentations or masking. 

Here, we follow a multimodal contrastive learning paradigm where (S+E) pairs are aligned during a pre-training stage. Specifically, we address the modality heterogeneity gap by employing \emph{modality-specific} encoders yielding a slide and expression embeddings that are aligned using a symmetric contrastive objective. Our models are based on large cohorts of publicly available (S+E) pairs, namely The Cancer Genome Atlas (TCGA) developed for studying human cancer and the Toxicogenomics Project-Genomics Assisted Toxicity Evaluation System (TG-GATEs) developed for assessing drug toxicity in rat model animals. (S+E) models are trained on multiple species (\textit{Homo sapiens} and \textit{Rattus norvegicus}) and sites (liver, breast, and lung), that we test on a panel of downstream tasks. To summarize, our contributions are: (1) the first self-supervised vision encoder for rat tissue trained on 15 million patches from 47,227 WSIs; (2) $\ours$, a transcriptomics-guided slide representation learning framework trained on thousands of (S+E) pairs using multimodal contrastive learning;
(3) a series of few-shot classification, prototype-based classification, and slide retrieval experiments for lesion classification in rat liver and cancer subtyping in human breast and lung cancer that show the predictive capabilities of $\ours$; and (4) a post-hoc interpretability analysis that enables deriving insights about the aligned latent space.

\begin{figure*}[t]
   \centering
   \includegraphics[width=0.9\linewidth]{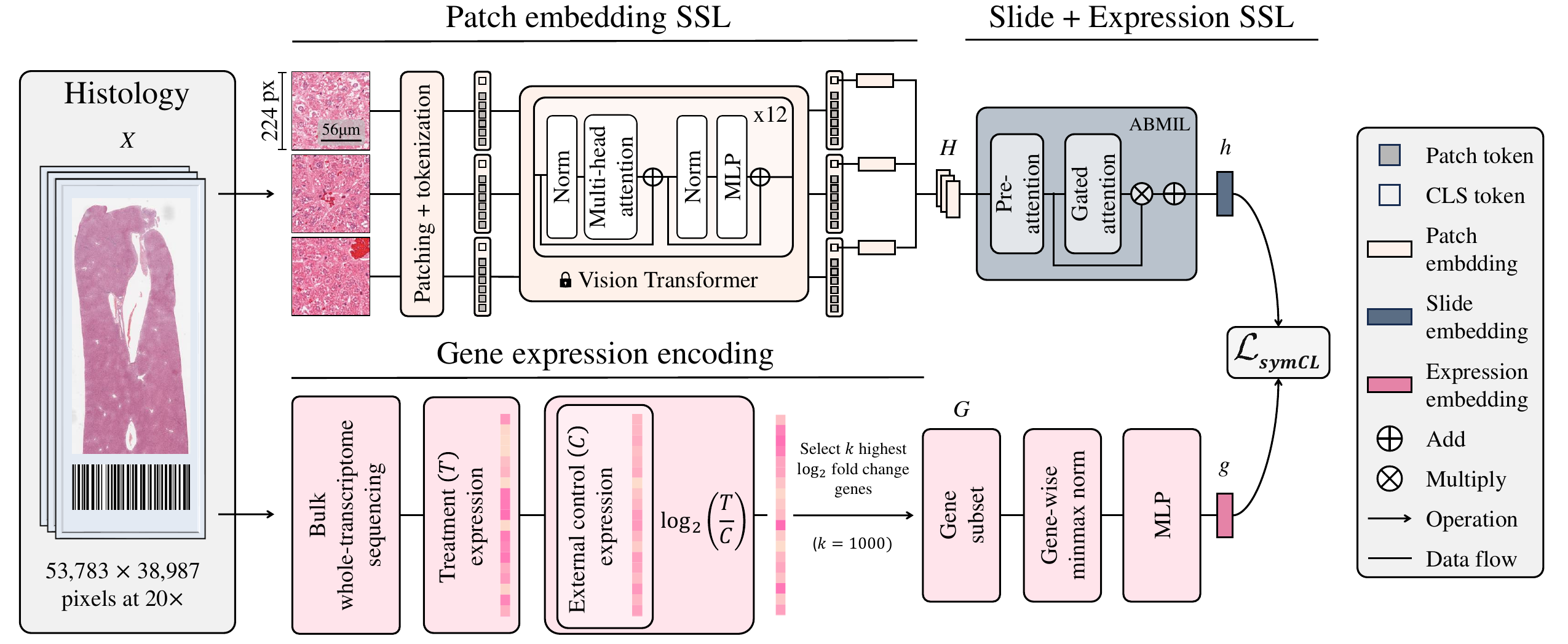}
   \caption{\textbf{Overview of $\ours$ for (S+E) pre-training}. An input histology slide is tessellated into patches and encoded using a pre-trained vision encoder. The resulting patch embeddings are passed to an ABMIL module to derive a slide embedding. The corresponding gene expression data are encoded using an MLP. A symmetric contrastive objective $\mathcal{L}_{symCL}$ learns to align embeddings from both modalities. During inference, a query slide is encoded into a slide embedding by the trained pooling module to be used for downstream tasks.}
   \label{fig:overview}
\end{figure*}

\section{Related work}
\label{sec:related_work}

\subsection{Self-supervised visual representation learning}

The combination of Vision Transformers (ViTs)~\cite{vaswani2017attention, dosovitskiy2020image} and SSL~\cite{zhou2022image,caron2021emerging} has proved to be a powerful tool for building task-agnostic image representations. SSL can be categorized into (1) contrastive approaches~\cite{radford2021learning,caron2021emerging}, whose underlying principle is to attract different representations of the same image (e.g., two distinct augmentations) while simultaneously pushing away representations of dissimilar images; (2) reconstruction approaches~\cite{he2022masked,xie2022simmim}, which aim to recover specific portions of an occluded image from the remaining parts of the same image; and (3) approaches combining both objectives~\cite{zhou2022image,oquab2023dinov2}. Representation learning in vision has also evolved towards multimodal vision-language models~\cite{radford2021learning,jia2021scaling,li2021align,singh2022flava,yu2022coca,alayrac2022flamingo,li2023blip,li2023scaling,wang2023image}. The same principles remain, where, for instance, the embedding of an image caption can be pulled close to the image (in a contrastive fashion), or partially masked with the objective to reconstruct the caption from the image. Vision-language models are also becoming prevalent in medical applications, by leveraging medical reports and textbooks~\cite{wang2022medclip,lin2023pmc}. Our work aligns with this idea, where we align expression profiles with the slide representation. 

\subsection{Self-supervised learning in CPath}

\textbf{Encoding histology patches:} Most works applying SSL to CPath focus on building embeddings from image patches (typically 256 $\times$ 256-pixel regions)~\cite{koohbanani2021self,filiot2023scaling,vorontsov2023virchow,wang2021transpath,kang2023benchmarking,ciga2022self,wang2022transformer,azizi2023robust,chen2024towards,lu2024towards}. State-of-the-art methods are using a combination of contrastive- and reconstruction-based objectives trained with a student-teacher learning paradigm~\cite{filiot2023scaling,vorontsov2023virchow}. Patch-level SSL is trained on increasingly large datasets and models (\emph{e.g.,} ViT-Huge trained on 1.5M slides in~\cite{vorontsov2023virchow}). These can be based on public archives such as TCGA or CPTAC~\cite{wang2021transpath,ciga2022self,wang2022transformer,azizi2023robust,filiot2023scaling}, on internal cohorts~\cite{vorontsov2023virchow}, or a mix of public and private datasets~\cite{kang2022benchmarking}. Recently, vision-language encoders designed for pathology have also been proposed~\cite{gamper2021multiple,lu2023visual,huang2023visual}, and rely on large-scale paired data scraped from social media, textbooks, or publications. All these models are solely based on human tissue, most of which are cancer samples. Here, we complement these by introducing the first vision encoder for rodent tissue microscopy, which plays a pivotal role in drug safety and biomarker discovery.

\textbf{Encoding histology slides:} Methods to build slide embeddings using SSL remain relatively scarce. Chen et al.~\cite{chen2022scaling} proposed a three-stage pre-training pipeline to hierarchically aggregate increasingly large tiles, from patches to patch embeddings to region embeddings to slide embeddings. Follow-up works improved pre-training using more complex training signals based on intra- and inter-slide similarity losses~\cite{yu2023slpd,lazard2023giga}, masked autoencoding~\cite{jiang2023masked} or patch prototyping~\cite{song2024morphological}.

\subsection{Supervised learning in CPath}

\textbf{Multiple Instance Learning:} MIL~\cite{dietterich1997solving} is the current \emph{de-facto} approach for WSI classification. In particular, Attention-based MIL (and its many extensions) has been used extensively in CPath~\cite{ilse2018attention,yao2021whole,lu2021data,zhan2022dtfdmil,yang2022concl,tu2022dual,qu2022bidirectional,wang2022sclwc,cui2023bayesmil,xiang2023exploring,javed2022additive,tang2023multiple,shao2023lnplmil,qu2023boosting,li2023task,lin2023interventional}. Context-aware extensions have also been proposed, such as based on graph neural networks~\cite{lee2022derivation,chan2023histopathology,nakhli2023copilot} and Transformers~\cite{shao2021transmil,myronenko2021accounting}. During (S+E) pre-training, we also employ MIL to pool pre-extracted patch embeddings into a slide embedding that we further use for SSL contrastive learning. 

\noindent\textbf{Multimodal learning:} While the representation learning capabilities of (S+E) pre-training remain poorly understood, the multimodal integration of histology with gene expression data has been extensively studied in cancer-specific and pan-cancer works, especially for prognostication~\cite{mobadersany2018predicting,chen2020pathomic,schmauch2020deep,wang2021GPDBN,ash2021joint,xie2023spatially,li2022hfbsurv,jaume2024modeling,chen2022pan}. Several mechanisms have been proposed such as late~\cite{chen2022pan} or early fusion using multimodal token aggregration~\cite{xu2023multimodal,zhou2023cross,jaume2024modeling}. Although not directly connected to our approach, they motivate exploring the connection between gene expression profiles and tissue morphology. Notably, recent studies more closely aligned with (S+E) pre-training and demonstrated improved multimodal downstream performance through multimodal pre-training utilizing histology and expression data~\cite{jin2023gene,zhou2023cross,ding2023pathology}.

\noindent\textbf{Computational Toxicologic Pathology (CompToxPath):} The majority of work in CPath is centered around studying human cancer. CompToxPath is emerging as a new sub-field that aims to augment drug safety assessment using AI, especially at the pre-clinical stage~\cite{mehrvar2021deep}. CompToxPath has been used for organ identification~\cite{citlalli2022mmonet}, detecting abnormalities ~\cite{holger2021histonet,baek2022artifical,shimazaki2022deep,hwang2023comparative}, such as necrosis and hypertrophy detection. However, none of these works include SSL or large-scale evaluations. This work bridges this gap by applying (S+E) pre-training to large-scale toxicology datasets.

\begin{figure*}[t]
   \centering
   \includegraphics[width=0.9\linewidth]{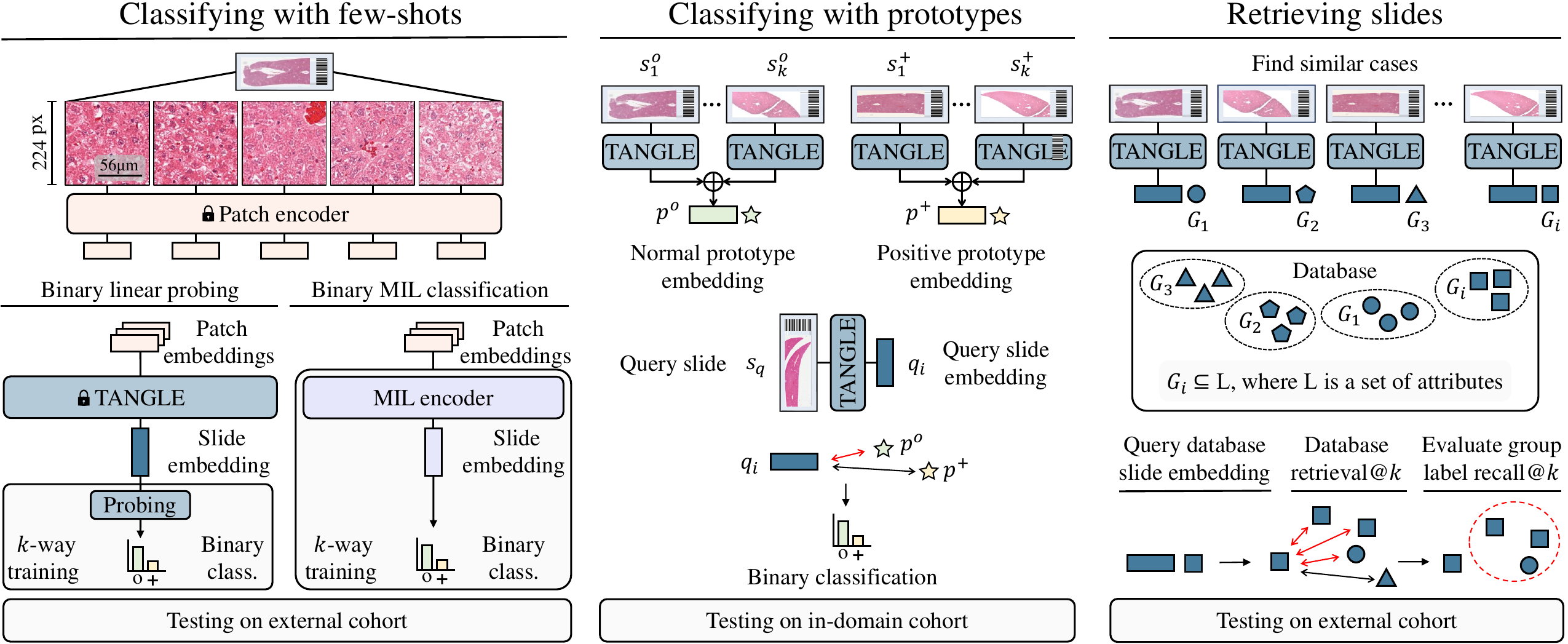}
   \caption{\textbf{Downstream tasks.} We test $\ours$ and baselines on (1) few-shot,(2) prototype-based classification, and (3) slide retrieval.
   }
   \label{fig:downstream}
\end{figure*}

\section{Method}
\label{sec:method}

Here, we present our framework, $\ours$, for TrANscriptomics-Guided sLidE representation learning (see Figure~\ref{fig:overview}). $\ours$ is composed of (1) a vision encoder that encodes patches into \emph{patch embeddings}, followed by a pooling module for learning a \emph{slide embedding} (Section~\ref{sec:patch_embedding}), (2) a gene expression encoder that combines transcriptomic measurements into an \emph{expression embedding} (Section~\ref{sec:expression_embedding}), and (3) a multimodal alignment module that learns to align both spaces (Section~\ref{sec:alignment}). $\ours$ is tested on a variety of downstream tasks (Section~\ref{sec:experiments}). 

\subsection{Slide encoder} \label{sec:patch_embedding}

Given a histology slide $\mathbf{X}_i \in \real^{d_x \times d_y \times 3}$, we follow the MIL paradigm~\cite{dietterich1997solving,ilse2018attention,lu2021data,shao2021transmil,lee2022derivation,li2021dual}, which consists of tessellating the slide into small patches, using a pre-trained vision encoder to extract patch embeddings, and pooling the resulting patch embeddings into a slide embedding. 

\noindent\textbf{Pre-trained patch encoding:} For encoding rat tissue, we trained from scratch a ViT-Base (86 million parameters) with iBOT~\cite{zhou2022image} on 15 million H\&E patches extracted from 47,227 WSIs for 80 epochs, which we denote as iBOT-Tox. This is, to date, the largest SSL model for non-human histology tissue (see \textbf{Supplemental}). For encoding human tissue, we use CTransPath~\cite{wang2021transpath,wang2022transformer}, a state-of-the-art publicly available vision encoder. CTransPath was trained on 15 million patches from over 32,000 WSIs using a tiny Swin Transformer~\cite{liu2021swin}. We denote the resulting patch embeddings of the $i$-th slide $\mathbf{X}_i$ as $\textbf{H}_{i} \in \real^{N_{\histology} \times d_{\histology}}$, where $N_{\histology}$ is the number of patch embeddings and $d_{\histology}$ their dimension.
 
\noindent\textbf{MIL slide encoding:} We learn a function $f(\mathbf{H}_{i}):\mathbb{R}^{N_{\histology} \times d_{\histology}} \rightarrow \real^{d}$ that maps the set of patch embeddings $\mathbf{H}_{i} \in \real^{N_{\histology} \times d_{\histology}}$ into a slide embedding $\mathbf{h}_{i} \in \real^d$. Here, we use the popular attention-based MIL model (ABMIL)~\cite{ilse2018attention}, which consists of learning patch-level attention weights used for pooling embeddings into a slide embedding.  
 
\subsection{Gene expression encoder} \label{sec:expression_embedding}

Given a set of raw transcriptomic measurements quantified across $N_{\genes}$ genes, we compute the log2 fold change relative to a control group, which represents gene expression deviations from a normal sample and, therefore, encode the magnitude of gene overexpression or underexpression (see \textbf{Supplemental}). The log2 fold change normalized transcriptomics associated with $\mathbf{X}_i$, denoted as $\textbf{t}_{i} \in \real^{N_{\genes}}$, can be seen as tabular data, which can efficiently be encoded with a multilayer perceptron (MLP) and named as $\phi(\cdot)$. Specifically, we train a 3-layer MLP to learn a mapping $\phi(\textbf{t}_{i}): \real^{N_{\genes}} \rightarrow \real^{d}$ to project a set of selected gene expressions $\textbf{t}_{i} \in \real^{N_{\genes}}$ to an expression embedding $\textbf{g}_{i}\in \real^d$. 

\subsection{Multimodal alignment} \label{sec:alignment}

\noindent\textbf{Pre-training contrastive alignment:} We align the embedding space of the slide and expression encoders using a symmetric cross-modal contrastive learning objective. This is a widely employed representation learning formulation~\cite{radford2021learning}, especially in visual-language pre-training~\cite{lu2023visual}. Formally, we define a batch as a set of $M$ (S+E) pairs $(\textbf{h}_i, \textbf{g}_i)_{i=1}^{M}$, where $\textbf{h}_i$ and $\textbf{g}_i$ are the $i$-th slide embedding and gene expression profiles, respectively. For a given pair ($\textbf{h}_i, \textbf{g}_i$), the objective is given by: 
\begin{equation}
\label{eq:loss_symcl}
\begin{split}
\mathcal{L}_{SymCL} = &-\frac{1}{2M}\sum_{i=1}^{M} \log \frac{\exp \left(\tau \boldsymbol{h}_i^{T} \boldsymbol{g}_i\right)}{\sum_{j=1}^{M} \exp \left(\tau \boldsymbol{h}_i^{T}  \boldsymbol{g}_j\right)} \\
& -\frac{1}{2M}\sum_{j=1}^{M} \log \frac{\exp \left(\tau \boldsymbol{g}_j^{T}  \boldsymbol{h}_j\right)}{\sum_{i=1}^{M} \exp \left(\tau \boldsymbol{g}_j^{T}  \boldsymbol{h}_i\right)} \\ 
\end{split}
\end{equation}
where the first term represents the slide-to-expression contrastive loss, and the second term represents the expression-to-slide contrastive loss. Each term maximizes the dot-product similarity between embeddings from the same pair normalized (with Softmax) by negative pairs, which can be interpreted as other ``classes".

\noindent\textbf{Complementary objective:} As an alternative to the contrastive loss, we introduce (1) an \emph{expression reconstruction} objective $\mathcal{L}_{\textsc{Rec}}$ framed as an expression regression task, and (2) a vision-only intra-modality objective $\mathcal{L}_{\textsc{Intra}}$ that aims to align different random subsets of the slide (local--local alignment) and random subsets with the average patch embedding (local--global alignment). We express these as,
\begin{equation}
\label{eq:rec}
\begin{split}
    \mathcal{L}_{\textsc{Rec}} =& \frac{1}{M} \sum_{i=1}^M \lVert \mathbf{g}_i - c\big(f(\mathbf{H}_i)\big) \rVert_2 \\
\end{split}
\end{equation}
\begin{equation}
\label{eq:intra}
\begin{split}
    \mathcal{L}_{\textsc{Intra}} =& -\frac{1}{2M} \sum_{i=1}^M  \log(\text{Softmax}({\mathbf{h}}_{i,1}^T\overline{\mathbf{h}}_i, \tau)  \\
    & -\frac{1}{2M} \sum_{i=1}^M  \log(\text{Softmax}({\mathbf{h}}_{i,1}^T{\mathbf{h}}_{i,2}, \tau) 
\end{split}
\end{equation}
where $c(\cdot)$ is an MLP regressor, $\overline{\mathbf{h}}_i$ is the average patch embedding $\overline{\mathbf{h}}_i = \frac{1}{N^{(i)}_{\histology}} \sum_j \mathbf{H}^{N^{(i)}_{\histology}}_{ij}$, and where ${\mathbf{h}}_{i,1}$ and ${\mathbf{h}}_{i,2}$ are slide embedding views derived from different random patch embedding subsets (\emph{e.g.,} 2048 patches). These variants are referred to as $\ours$-\textsc{Rec} and $\textsc{Intra}$, respectively. 

\noindent\textbf{Inference:} During inference, the query slide is passed through the vision encoder to extract patch embeddings and then to the MIL module to derive the slide embedding that encodes the morphological manifestations of the corresponding molecular signatures. We use the resulting slide embeddings for few-shot classification using linear probing and prototyping, and slide retrieval (see Figure~\ref{fig:downstream}).

\begin{table*}
\centering
\caption{\textbf{Few-shot lesion classification in rat liver.} Comparison of lesion classification (multi-label classification) using MIL \emph{vs.} $\ours$ and variations with linear probing, and evaluated using Macro-AUC (as \%). All models are tested on an independent test cohort comprising 4,584 slides, without any data leakage from unimodal and multimodal pre-training. Standard deviation reported over five runs.}
\label{tab:few_shot_tggates}
\scalebox{0.9}{
\begin{tabular}{ll|ccccc}
\toprule
& Model/Data & $k$=1$(\uparrow)$ & $k$=5$(\uparrow)$ & $k$=10$(\uparrow)$ & $k$=25$(\uparrow)$ & $k$=50$(\uparrow)$\footnotemark \\
\midrule
\parbox[t]{0mm}{\multirow{6}{*}{\rotatebox[origin=c]{90}{{MIL}}}} 
& ResNet50+TransMIL~\cite{shao2021transmil}     & 53.3 $\pm$ 3.1 & 48.2 $\pm$ 2.9 & 53.2 $\pm$ 2.3 & 52.5 $\pm$ 3.7 & 52.9 $\pm$ 4.2 \\
& CTransPath+TransMIL~\cite{shao2021transmil}      & 50.1 $\pm$ 4.1 & 51.1 $\pm$ 0.8 & 55.4 $\pm$ 3.9 & 58.1 $\pm$ 3.8 & 65.9 $\pm$ 4.2 \\
& iBOT-Tox+TransMIL~\cite{shao2021transmil}          & 55.6 $\pm$ 6.1 & 66.5 $\pm$ 6.4 & 66.3 $\pm$ 6.2 & 68.6 $\pm$ 9.8 & 70.4 $\pm$ 10.6 \\
& ResNet50+ABMIL~\cite{ilse2018attention}     & 56.0 $\pm$ 4.5 & 59.1 $\pm$ 7.1 & 64.1 $\pm$ 5.9 & 74.2 $\pm$ 8.6 & 80.3 $\pm$ 5.8 \\
& CTransPath+ABMIL~\cite{ilse2018attention}     & 59.5 $\pm$ 4.4 & 71.7 $\pm$ 8.0 & 73.8 $\pm$ 9.5 & 78.7 $\pm$ 9.4 & 81.0 $\pm$ 7.3 \\
& iBOT-Tox+ABMIL~\cite{ilse2018attention}          & 61.7 $\pm$ 5.3 & 73.2 $\pm$ 6.8 & 78.8 $\pm$ 9.3 & 81.6 $\pm$ 6.9 & 83.8 $\pm$ 8.1 \\
\midrule
\parbox[t]{0mm}{\multirow{6}{*}{\rotatebox[origin=c]{90}{{Linear probing}}}} & ResNet50+Avg.~\cite{he2016deep}     & 55.0 $\pm$ 3.3 & 57.7 $\pm$ 11.8 & 60.5 $\pm$ 9.6 & 68.6 $\pm$ 8.0 & 72.7 $\pm$ 7.8 \\
& CTransPath+Avg.~\cite{wang2022transformer}   & 56.9 $\pm$ 4.4 & 56.5 $\pm$ 10.5 & 61.9 $\pm$ 8.3 & 70.5 $\pm$ 8.1 & 73.9 $\pm$ 6.1 \\
& iBOT-Tox+Avg. (ours)      & 53.9 $\pm$ 5.3 & 63.5 $\pm$ 6.9 & 71.5 $\pm$ 6.1 & 79.7 $\pm$ 5.0 & 81.9 $\pm$ 6.2 \\
& iBOT-Tox+Intra (ours)       & 56.3 $\pm$ 7.3 & 62.6 $\pm$ 10.3 & 72.7 $\pm$ 7.4 & 80.2 $\pm$ 8.4 & 83.3 $\pm$ 8.0 \\
& $\ours$-Rec (ours) & \textbf{73.8} $\pm$ 13.5 & 75.5 $\pm$ 14.3 & 78.3 $\pm$ 12.2 & 81.8 $\pm$ 10.8 & 82.7 $\pm$ 8.8 \\
& $\ours$ (ours)          & 72.1 $\pm$ 11.6 & \textbf{80.1} $\pm$ 11.3 & \textbf{84.7} $\pm$ 9.0 & \textbf{86.3} $\pm$ 7.9 & \textbf{86.9} $\pm$ 7.6 \\
\bottomrule
\end{tabular}
}
\end{table*}
\footnotetext{50 or maximal available labeled samples per class}

\section{Experiments and results}
\label{sec:experiments}

\subsection{Dataset} 

\textbf{TG-GATEs:} We collected all slides from the TG-GATEs portal~\cite{igarashi2014open}, which comprises 23,136 liver and 28,747 kidney slides ($\approx$ 25TB of raw data). All slides are liver and kidney sections from Sprague-Dawley (SD) rats acquired in pre-clinical drug safety studies on 157 compounds. Each slide represents the morphological changes (lesions) observed after the administration of a particular drug dosage at a specified time point of sacrifice, denoted as a \emph{sample group}. We manually curated the liver annotations into six classes (multi-label classification). We used a subset of 29 studies (n=4,584 WSIs, liver only) as an independent test cohort. Other studies (both liver and kidney slides) are used for iBOT-Tox pre-training, (S+E) pre-training, and few-shot training. We additionally collected the corresponding gene expression profiles (microarrays) of 6,597 liver slides and selected the top 1,000 genes with the largest log2 fold change (see \textbf{Supplemental}).  

\noindent\textbf{TCGA:} We collected 1,041 primary cases from the TCGA Breast Invasive Carcinoma (TCGA-BRCA) cohort, which comprises 831 Invasive Ductal Carcinoma (IDC) and 210 Invasive Lobular Carcinoma (ILC). We additionally collected 1,031 primary cases from the TCGA Non-Small Cell Lung Cancer (TCGA-NSCLC) cohort, among which 528 cases of Lung Adenocarcinoma (TCGA-LUAD) and 505 cases of Lung Squamous Cell Carcinoma (TCGA-LUSC). For each case, we downloaded the corresponding gene expression data (RNA sequencing) from the Xena database~\cite{goldman2020visualizing} that we curated using the method in~\cite{jaume2024modeling}, resulting in 4,999 gene expression per case.

\noindent\textbf{In-house:} We also collected a BRCA (n=1,265 slides, 982 IDC and 283 ILC) and NSCLC (n=1,946 slides, n=1,621 LUAD and n=325 LUSC) cohort from our in-house archives. These two cohorts are used as independent test sets for which gene expression data are not required. Slides from all datasets were processed at 20$\times$ magnification (0.5$\mu$m/px).

\subsection{Linear probing few-shot classification}

\begin{table*}
\centering
\caption{\textbf{Few-shot cancer subtype classification in human breast and lung.}
All models are tested on an independent test cohort comprising 1,265 breast slides and 1,946 lung slides and evaluated using Macro-AUC. Standard deviation reported over ten runs.}
\label{tab:brca_fewshot}
\scalebox{0.9}{
\begin{tabular}{ll|cccc|cccc}
\toprule
& Model/Data & \multicolumn{4}{c}{\textbf{Breast}} & \multicolumn{4}{c}{\textbf{Lung}} \\
& & $k$=1$(\uparrow)$ & $k$=5$(\uparrow)$ & $k$=10$(\uparrow)$ & $k$=25$(\uparrow)$ &  $k$=1$(\uparrow)$ & $k$=5$(\uparrow)$ & $k$=10$(\uparrow)$ & $k$=25$(\uparrow)$ \\
\midrule
\parbox[t]{0mm}{\multirow{8}{*}{\rotatebox[origin=c]{90}{MIL}}} 
& \multirow{1}{*}{ResNet50+TransMIL~\cite{shao2021transmil}}     & 49.4 & 50.5  & 53.7  & 51.8  & 55.9  & 55.0  & 54.2  & 52.8   \\
& & {\footnotesize(}${\scriptstyle\pm}$ {\footnotesize 13.0)}& {\footnotesize(}${\scriptstyle\pm}$ {\footnotesize 7.6)}  & {\footnotesize(}${\scriptstyle\pm}$ {\footnotesize 8.8)}  & {\footnotesize(}${\scriptstyle\pm}$ {\footnotesize 4.9)}  & {\footnotesize(}${\scriptstyle\pm}$ {\footnotesize 5.4)} & {\footnotesize(}${\scriptstyle\pm}$ {\footnotesize 5.6)} & {\footnotesize(}${\scriptstyle\pm}$ {\footnotesize 6.1)} & {\footnotesize(}${\scriptstyle\pm}$ {\footnotesize 5.4)}  \\

& \multirow{1}{*}{CTransPath+TransMIL~\cite{shao2021transmil}}     & 55.5 & 63.0 & 63.9 & 71.2 & 54.1 & 64.8 & 68.4 & 80.5  \\
& & {\footnotesize(}${\scriptstyle\pm}$ {\footnotesize 9.5)} & {\footnotesize(}${\scriptstyle\pm}$ {\footnotesize 9.1)}  & {\footnotesize(}${\scriptstyle\pm}$ {\footnotesize 7.8)}  & {\footnotesize(}${\scriptstyle\pm}$ {\footnotesize 12.7)}  & {\footnotesize(}${\scriptstyle\pm}$ {\footnotesize 8.6)} & {\footnotesize(}${\scriptstyle\pm}$ {\footnotesize 8.9)} & {\footnotesize(}${\scriptstyle\pm}$ {\footnotesize 10.4)} & {\footnotesize(}${\scriptstyle\pm}$ {\footnotesize 10.8)}  \\

& \multirow{1}{*}{ResNet50+ABMIL~\cite{ilse2018attention}}     & 53.9  & 58.0  & 67.6  & 71.0 &  58.2 & 65.9  & 65.6  & 64.8  \\
& & {\footnotesize(}${\scriptstyle\pm}$ {\footnotesize 14.4)} & {\footnotesize(}${\scriptstyle\pm}$ {\footnotesize 9.9)}  & {\footnotesize(}${\scriptstyle\pm}$ {\footnotesize 9.6)}  & {\footnotesize(}${\scriptstyle\pm}$ {\footnotesize 3.7)}  & {\footnotesize(}${\scriptstyle\pm}$ {\footnotesize 7.4)} & {\footnotesize(}${\scriptstyle\pm}$ {\footnotesize 6.1)} & {\footnotesize(}${\scriptstyle\pm}$ {\footnotesize 4.6)} & {\footnotesize(}${\scriptstyle\pm}$ {\footnotesize 1.4)}  \\

& \multirow{1}{*}{CTransPath+ABMIL~\cite{ilse2018attention}}     & 57.4 & 70.9  & 73.8  & 83.5 & 62.8  & 78.7  & 85.3  & 87.2 \\
& & {\footnotesize(}${\scriptstyle\pm}$ {\footnotesize 14.0)} & {\footnotesize(}${\scriptstyle\pm}$ {\footnotesize 10.5)}  & {\footnotesize(}${\scriptstyle\pm}$ {\footnotesize 7.1)}  & {\footnotesize(}${\scriptstyle\pm}$ {\footnotesize 8.6)}  & {\footnotesize(}${\scriptstyle\pm}$ {\footnotesize 9.0)} & {\footnotesize(}${\scriptstyle\pm}$ {\footnotesize 11.7)} & {\footnotesize(}${\scriptstyle\pm}$ {\footnotesize 4.5)} & {\footnotesize(}${\scriptstyle\pm}$ {\footnotesize 3.4)}  \\

\midrule
\parbox[t]{0mm}{\multirow{12}{*}{\rotatebox[origin=c]{90}{Linear probing}}} & \multirow{1}{*}{ResNet50+Avg.~\cite{he2016deep}} & 65.7 & 67.4  & 68.0 & 76.6 & 57.4 & 60.1 & 60.7 & 59.5 \\
& & {\footnotesize(}${\scriptstyle\pm}$ {\footnotesize 17.3)} & {\footnotesize(}${\scriptstyle\pm}$ {\footnotesize 13.1)}  & {\footnotesize(}${\scriptstyle\pm}$ {\footnotesize 13.9)}  & {\footnotesize(}${\scriptstyle\pm}$ {\footnotesize 8.0)}  & {\footnotesize(}${\scriptstyle\pm}$ {\footnotesize 6.5)} & {\footnotesize(}${\scriptstyle\pm}$ {\footnotesize 4.7)} & {\footnotesize(}${\scriptstyle\pm}$ {\footnotesize 4.2)} & {\footnotesize(}${\scriptstyle\pm}$ {\footnotesize 2.1)}  \\

& \multirow{1}{*}{CTransPath+Avg.~\cite{wang2022transformer}} & \textbf{68.6} & 71.3 & 71.3 & 80.0 & 58.2 & 66.0 & 71.0 & 75.2 \\
& & {\footnotesize(}${\scriptstyle\pm}$ {\footnotesize 16.9)} & {\footnotesize(}${\scriptstyle\pm}$ {\footnotesize 11.1)}  & {\footnotesize(}${\scriptstyle\pm}$ {\footnotesize 14.4)}  & {\footnotesize(}${\scriptstyle\pm}$ {\footnotesize 7.5)}  & {\footnotesize(}${\scriptstyle\pm}$ {\footnotesize 6.6)} & {\footnotesize(}${\scriptstyle\pm}$ {\footnotesize 6.6)} & {\footnotesize(}${\scriptstyle\pm}$ {\footnotesize 2.6)} & {\footnotesize(}${\scriptstyle\pm}$ {\footnotesize 3.3)}  \\

& \multirow{1}{*}{$\text{HIPT}_{\text{CLS-4K}}$~\cite{chen2022scaling}} & 62.2 & 63.7 & 71.0 & 78.1 & 59.8 & 70.5 & 74.1 & 79.1\\
& & {\footnotesize(}${\scriptstyle\pm}$ {\footnotesize 10.3)} & {\footnotesize(}${\scriptstyle\pm}$ {\footnotesize 11.6)}  & {\footnotesize(}${\scriptstyle\pm}$ {\footnotesize 11.1)}  & {\footnotesize(}${\scriptstyle\pm}$ {\footnotesize 6.2)}  & {\footnotesize(}${\scriptstyle\pm}$ {\footnotesize 6.5)} & {\footnotesize(}${\scriptstyle\pm}$ {\footnotesize 6.6)} & {\footnotesize(}${\scriptstyle\pm}$ {\footnotesize 3.4)} & {\footnotesize(}${\scriptstyle\pm}$ {\footnotesize 4.1)}  \\


& \multirow{1}{*}{CTransPath+Intra (ours)} & 57.2 & 73.2 & 71.9 & 83.2 & 59.6 & 70.3 & 75.4 & 83.2\\
& & {\footnotesize(}${\scriptstyle\pm}$ {\footnotesize 14.7)} & {\footnotesize(}${\scriptstyle\pm}$ {\footnotesize 5.5)}  & {\footnotesize(}${\scriptstyle\pm}$ {\footnotesize 9.1)}  & {\footnotesize(}${\scriptstyle\pm}$ {\footnotesize 6.8)}  & {\footnotesize(}${\scriptstyle\pm}$ {\footnotesize 7.0)} & {\footnotesize(}${\scriptstyle\pm}$ {\footnotesize 9.8)} & {\footnotesize(}${\scriptstyle\pm}$ {\footnotesize 6.7)} & {\footnotesize(}${\scriptstyle\pm}$ {\footnotesize 4.4)}  \\

& \multirow{1}{*}{$\ours$-Rec (ours)} & 56.3 & 73.6 & 68.3 & 83.4  & \textbf{81.6} & 84.1 & 85.5 & 86.6 \\
& & {\footnotesize(}${\scriptstyle\pm}$ {\footnotesize 19.6)} & {\footnotesize(}${\scriptstyle\pm}$ {\footnotesize 6.8)}  & {\footnotesize(}${\scriptstyle\pm}$ {\footnotesize 10.1)}  & {\footnotesize(}${\scriptstyle\pm}$ {\footnotesize 6.6)}  & {\footnotesize(}${\scriptstyle\pm}$ {\footnotesize 10.3)} & {\footnotesize(}${\scriptstyle\pm}$ {\footnotesize 4.9)} & {\footnotesize(}${\scriptstyle\pm}$ {\footnotesize 1.8)} & {\footnotesize(}${\scriptstyle\pm}$ {\footnotesize 2.3)}  \\

& \multirow{1}{*}{$\ours$ (ours)}  & 67.3 & \textbf{82.6} & \textbf{84.8} & \textbf{90.3} & 70.9 &  \textbf{89.3} & \textbf{91.5} & \textbf{93.9} \\
& & {\footnotesize(}${\scriptstyle\pm}$ {\footnotesize 19.1)} & {\footnotesize(}${\scriptstyle\pm}$ {\footnotesize 8.0)}  & {\footnotesize(}${\scriptstyle\pm}$ {\footnotesize 5.0)}  & {\footnotesize(}${\scriptstyle\pm}$ {\footnotesize 3.7)}  & {\footnotesize(}${\scriptstyle\pm}$ {\footnotesize 6.0)} & {\footnotesize(}${\scriptstyle\pm}$ {\footnotesize 4.1)} & {\footnotesize(}${\scriptstyle\pm}$ {\footnotesize 2.1)} & {\footnotesize(}${\scriptstyle\pm}$ {\footnotesize 1.3)}  \\

\bottomrule
\end{tabular}
}
\end{table*}

We evaluate (S+E) pre-training in a few-shot classification scenario for lesion detection in liver (Table~\ref{tab:few_shot_tggates}), and breast and lung cancer subtyping (Table~\ref{tab:brca_fewshot}). Following standard practice in SSL~\cite{caron2021emerging,zhou2022image}, we employ linear probing for benchmarking $\ours$, $\ours$-$\textsc{Rec}$, and $\textsc{Intra}$. In addition, we benchmark HIPT~\cite{chen2022scaling} and baselines based on the average patch embeddings using different backbones (denoted as ResNet50+Avg., CTransPath+Avg. and iBOT-Tox+Avg.). Finally, we include two supervised MIL baselines, ABMIL~\cite{ilse2018attention} and TransMIL~\cite{shao2021transmil} (see Figure~\ref{fig:downstream}, left). Baselines are trained five times (Table~\ref{tab:few_shot_tggates}) and ten times (Table~\ref{tab:brca_fewshot}), using $k$ randomly sampled examples per class. 

\noindent\textbf{$\ours$ \emph{vs.} supervised MIL:} $\ours$ significantly outperforms all MIL baselines in the three datasets with an absolute gain of $+ 5.9\%$ in liver, $+ 11.0\%$ in breast, and $+ 6.2\%$ in lung compared to ABMIL for $k$=10. ABMIL leads to consistently better performance than TransMIL, which we hypothesize is due to (1) the use of a simpler architecture beneficial in low-data regimes and (2) tasks where the cellular morphology is more informative than the global tissue structure. 

\noindent\textbf{$\ours$ \emph{vs.} averaging \emph{vs.} MIL:} Despite the simplicity of these baselines, averaging provides performance that is on par with MIL in breast subtyping and liver lesion detection. We also observe that employing domain-specific vision encoders leads to substantial improvements, with CTransPath+Avg. outperforming ResNet50+Avg., which our iBOT-Tox+Avg. model in liver lesion detection significantly outperforms in TG-GATEs ($+9.6\%$ and $+11.0\%$ compared to CTransPath+Avg. and ResNet50+Avg. for $k$=10). 

\noindent\textbf{$\ours$ \emph{vs.} $\textsc{Intra}$ \emph{vs.} HIPT:} $\textsc{Intra}$ and HIPT provide similar performance in breast and lung, but are both significantly outperformed by $\ours$ ($+12.9\%$ and $+16.1\%$ for $k$=10 in breast and lung compared to $\textsc{Intra}$). Both HIPT and $\textsc{Intra}$ are only marginally better or similar to the average patch embedding, which highlights the complexity of slide-level SSL. 

\noindent\textbf{$\ours$ \emph{vs.} $\ours$-$\textsc{Rec}$:} $\ours$-$\textsc{Rec}$ shows surprisingly high performance for $k$=1, but is outperformed for larger values of $k$. We hypothesize that $\ours$-$\textsc{Rec}$ renders simplified embeddings (\emph{i.e.,} low-rank, see next Section), which makes one-shot learning easier but cannot express complex morphological subtleties. 

\noindent\textbf{Loss ablation:} In TG-GATEs relative to
$\ours$, adding a $\ours$-$\textsc{Rec}$ objective gives +0.05\% AUC, adding $\textsc{Intra}$ on top gives -0.8\% AUC, and -2.0\% AUC when solely complementing $\ours$ with $\textsc{Intra}$. We hypothesize that staining differences between train and test cause the \textsc{Intra} objective to overfit, leading to worse performance. Replacing the cross-modal contrastive loss with an L1 objective gives -6.7\% AUC and -7.0\% AUC with an L2 (some designs conceptually similar to~\cite{zhou2023cross,ding2023pathology,jin2023gene}, see \textbf{Supplemental}). 

\subsection{Prototyping few-shot classification}

We also assess the capacity of $\ours$ to construct slide-level prototypes capable of predicting specific morphological characteristics. Specifically, we define a positive slide prototype $p^{+}$ as the average of $k$ ($k$=1,3,5) slide embeddings with a morphology of interest. Similarly, a normal prototype $p^{0}$ is defined using $k$ normal slides, where the morphology under consideration is absent. Subsequently, we gauge the similarity between a query slide $q_i$ and the two prototypes using the L2-distance -- the distances interpreted as confidence prediction used for classification, \emph{i.e.,} $\lVert q_i-p^{+} \rVert$ and $\lVert q_i-p^{0} \rVert$, (see Figure~\ref{fig:downstream}, center). We apply this method to detect two types of lesions within the TG-GATEs test set, namely (1) eosinophilic degeneration in thioacetamide (n=170), and (2) bile duct proliferation in methylene dianiline (n=170). This setup mirrors a realistic application of AI, where the identification of a drug-induced morphology on $k$ slides enables detecting if this morphology is present in slides from the same study, thereby enabling synergies between doctors and AI systems.  

\begin{figure}[t]
   \includegraphics[width=\linewidth]{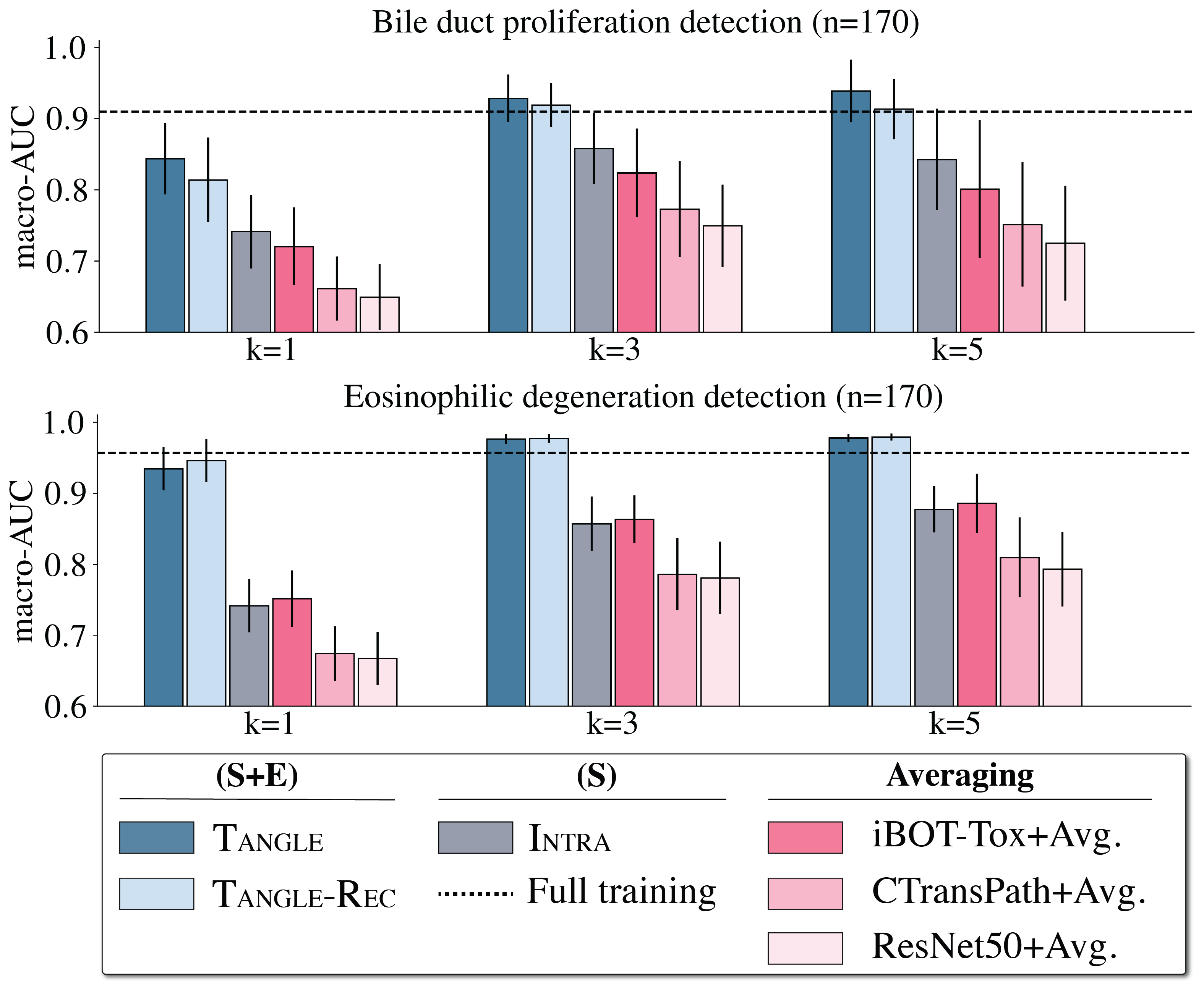}
   \caption{\textbf{Prototype-based classification.} Comparison of $\ours$ and baselines for identifying study-level morphologies evaluated using macro-AUC. Prototypes are defined as the average of $k$ slides selected from the study. Full training is an ABMIL trained on TG-GATEs train set (n=18,552). Standard deviation reported over 100 bootstrapping iterations.}
   \label{fig:prototype_classification}
\end{figure}

As shown in Figure~\ref{fig:prototype_classification}, $\ours$ and $\ours$-$\textsc{Rec}$ outperform all baselines in both studies. Compared to an ABMIL model trained on 100\% of TG-GATEs (n=18,552), $\ours$ with $k \geq 3$ leads to better performance. This highlights that (1) TG-GATEs includes study-specific morphologies that can be challenging to model, and (2) prototyping can help address this gap with minimal effort. 

\subsection{Slide retrieval}

We further evaluate $\ours$ on slide retrieval using TG-GATEs test set. Each slide is associated with four others that share the same sample group.
We extracted a subset of 594 slides with known drug-induced lesions. Our task is to retrieve all slides that share the same sample group characteristics as the query, thereby demonstrating the capability of $\ours$ to capture compound-, dose- and sacrifice-specific features. Specifically, we compute the Recall@$k$ ($k$=5, 10, 20), which measures the proportion of relevant slides that appear among the $k$ most similar slides, with four being the total number of slides to retrieve in this context. The slide similarity is quantified using the cosine distance metric applied to the unnormalized slide embeddings (see Figure~\ref{fig:downstream}, right).

\begin{figure}[t]
   \includegraphics[width=\linewidth]{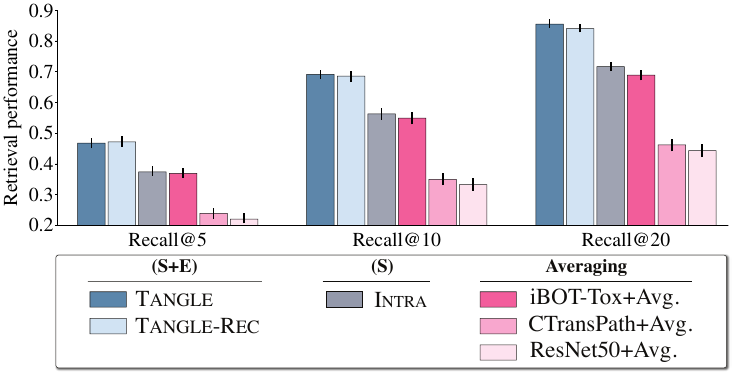}
   \caption{\textbf{Slide retrieval.} Comparison of $\ours$ and baselines for retrieving slides with drug-induced lesions from the same \emph{sample group} in TG-GATEs test. Recall@$k$ quantifies the count of retrieved instances within the top-$k$ most similar slides normalized by the number of instances to retrieve (four per \emph{sample group}). Standard deviation reported over 100 bootstrapping iterations.}
   \label{fig:slide_retrieval}
\end{figure}

As presented in Figure~\ref{fig:slide_retrieval}, $\ours$ reaches the best retrieval performance with on average 2.88/4 slides correctly retrieved among the top-$k$=10 instances and 3.44/4 among the top-$k$=20 instances. This result highlights that $\ours$ can capture subtle morphological differences, such as those induced by administering different doses or sacrifice times. 

Overall, results from Figure~\ref{fig:prototype_classification} and~\ref{fig:slide_retrieval} ascertain the conclusion from the few-shot evaluation in that (1) (S+E) pre-training can capture task-agnostic features that can be used for downstream tasks, (2) intra-modality pre-training can outperform averaging, but their training signal remains weak, and (3) in-domain patch feature extractors greatly improve downstream performance. Additional experiments ablating $\ours$ and $\textsc{Intra}$ losses, and showing the impact of hyper-parameters (batch size, temperature, number of sampled patches) are presented in the \textbf{Supplemental}.

\subsection{Interpretability}

\begin{figure*}[t]
   \centering
   \includegraphics[width=0.92\linewidth]{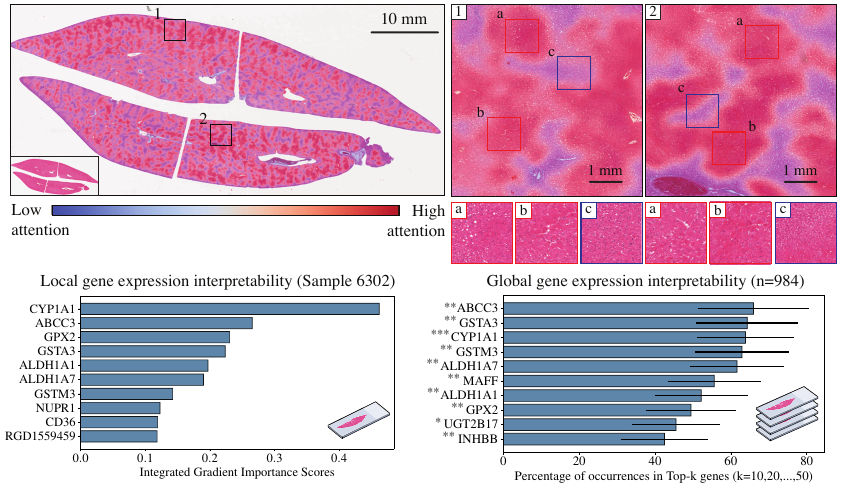}
   \caption{\textbf{Interpretability of $\ours$.} \textbf{Top:} Visualization of the attention weights of $\ours$ in a TG-GATEs liver slide. High-attention regions highlight lesions (hepatocellular hypertrophy and fatty change). \textbf{Left:} Integrated Gradient (IG) scores of the gene expression encoder. High-importance genes map to well-known markers of liver toxicity, such as CYP1A1. \textbf{Right:} Percentage of occurrence of the top-$k$ genes in test. Many genes consistently appear as influential ($>$40\% of tok-$k$ genes). * denotes the number of publications referencing this gene as connected to drug-induced liver
   injury according to the CTD database (*: $>$500, **: $>$1,000, ***: $>$2,000).
   }
   \label{fig:interpretability}
\end{figure*}

To better understand $\ours$ properties, we analyzed the \emph{rank} of the space spawned by the test slide embeddings (computed using the entropy of the $d$ largest singular values of the embedding matrix, see \textbf{Supplemental}). Indeed, \emph{rank} has been shown to be a predictor of downstream performance -- and constitute a necessary (but not sufficient) condition for discriminative latent spaces~\cite{garrido2022rankme}. We observe a strong positive correlation between rank and few-shot performance in all datasets among methods of the same family, (S+E), (S), and Averaging, as exemplified with $k$=10 (see \textbf{Supplemental}). This confirms the importance of building domain-specific feature encoders for increased expressivity. This also suggests that reconstruction-based methods suffer from some degree of dimensionality collapse, which we hypothesize stems from over-fitting (and might disappear with larger cohorts). Finally, $\textsc{Intra}$ models have high ranks despite performing significantly worse than (S+E), which might be explained by the latent space expressing clinically irrelevant factors, such as staining variations.

Furthermore, we investigated whether salient histologic and expression features align with previously established biological findings. First, we visualized the attention weights learned during $\ours$ pre-training (Figure~\ref{fig:interpretability}, top). Important regions with high attention (visible in red) correlate with lesions (fatty change and hepatocellular hypertrophy, see \textbf{Supplemental} for heatmaps of lung and breast cancer samples). Next, we applied Integrated Gradients (IG) to derive gene-level importance scores (Figure~\ref{fig:interpretability}, left) on TG-GATEs test samples with reported lesions. From there, we identified genes that consistently appear in the top-$k$ most influential genes, such as ABBC3 and CYPP1A1 (Figure~\ref{fig:interpretability}, right). We then quantitatively assessed their relevance by querying the Comparative Toxicogenomics Database (CTD) that aggregates all the literature on toxicology. 9/10 of the most important genes have more than 1,000 references connecting them to drug-induced liver injury, highlighting their relevance for slide representation learning.   

\section{Conclusion}

In this paper, we introduced Slide+Expression (S+E) pre-training for slide representation learning. Our approach, $\ours$, was trained and tested on several species (\textit{Homo sapiens} and \textit{Rattus norvegicus}) and tissue sites (breast, lung, and liver). Overall, $\ours$ outperforms all baselines significantly on several downstream tasks, including few-shot classification, prototype-based classification, and slide retrieval. These results highlight the potential of (S+E) pre-training and pave the way for additional developments~\cite{rahaman2023breast}. Future work includes exploring multimodal SSL objectives that extend beyond or synergize with, contrastive approaches, such as reconstruction of multimodal masks. Concurrently, evaluating (S+E) pre-training on more tasks, such as predicting hormone receptor status from H\&E slides, are promising research directions. 

\clearpage

{
    \small
    \bibliographystyle{ieeenat_fullname}
    \bibliography{main}
}

\clearpage
\clearpage
\setcounter{page}{1}
\setcounter{section}{0}
\setcounter{figure}{0}
\setcounter{table}{0}
\setcounter{equation}{0}

\renewcommand{\thesection}{\Alph{section}}
\renewcommand{\tablename}{Supplemental Table}
\renewcommand{\figurename}{Supplemental Figure}

\maketitlesupplementary

\section{Model \& training}

\noindent\textbf{iBOT-Tox pre-training:} iBOT-Tox is the first vision encoder for toxicologic pathology targeting non-human samples. It uses a Vision Transformer Base (ViT-B)~\cite{dosovitskiy2020image} as backbone to learn 768-dimensional embeddings from 224$\times$224 pixels image patches. ViTs are based on the self-attention paradigm to encode spatial interactions among small regions (called tokens) of the input image. iBOT-Tox is trained using the iBOT recipe~\cite{zhou2022image}, a state-of-the-art training strategy based on student-teacher knowledge distillation~\cite{caron2021emerging}. iBOT combines contrastive and reconstruction objectives: (1) a self-distillation objective to align different views of the input image based on image crop and augmentation. This objective helps to encode contextual and semantic information from the image, allowing for the creation of representations that are invariant to staining or rotation; and (2) a masked image modeling objective that aims to reconstruct image tokens from the other tokens. This objective helps to encode the image structure and is analogous to masked language modeling in Large Language Model training, such as BERT\cite{devlin2018bert}. To train the network, we relied on the public implementation of iBOT\footnote{https://github.com/bytedance/ibot}. iBOT-Tox was trained on 15 million patches extracted from different 47,227 WSIs (liver and kidney slides). We trained the network for 1,176,640 iterations or 80 epochs. The specific hyperparameters used for training are listed in Table~\ref{tab:hparams_ibot}. Most parameters were adapted from ImageNet-22K pre-training.

\noindent\textbf{ABMIL architecture:} $\ours$ is using an ABMIL architecture~\cite{ilse2018attention,lu2021data}, which is composed of three components: a pre-attention MLP, consisting of 2 layers with 768 hidden units, layer normalization, GELU activation, and 0.1 dropout; a gated-attention network, consisting of 2-layer MLP with 512 hidden units, with Sigmoid and Tanh activation respectively and 0.25 dropout; and a post-attention network, consisting a linear layer with 768 units.

\noindent\textbf{$\ours$ pre-training:} We pre-trained $\ours$ with AdamW optimizer and a batch size of 128 for 50 epochs. The learning rate is linearly ramped up during a 5-epoch warmup from 1e-8 to 1e-4. Then, we employed a cosine scheduler to reach the final learning rate of 1e-8 after 50 epochs. To increase training diversity and simplify batch processing, we sample a fixed and random subset of patches per slide. In TG-GATEs, we sample 4,096 patches, and in TCGA-BRCA and TCGA-NSCLC, we sample 2,048 patches per slide. In slides with fewer patches, we perform random over-sampling. 

\section{Data}

\textbf{TG-GATEs transcriptomics pre-processing:} The raw transcriptomics consists of microarrays (Affymetrix GeneChip) with 31,042 probes. Data were downloaded from the toxigates portal\footnote[1]{https://toxygates.nibiohn.go.jp/toxygates/} that aggregates all omics data acquired as part of The Japanese Toxicogenomics Project~\cite{igarashi2014open}. All data followed probe-wise normalization using log2 fold change with respect to a control group. Log2 fold change quantifies the proportional difference, on a logarithmic scale, between the expression levels of a particular probe under two conditions: a control group (on average 22 slides per study in TG-GATEs) and a sample group (a defined set of compound, time and sacrifice). Each probe was then mapped to a unique gene identifier using SynGoPortal\footnote[2]{https://www.syngoportal.org/convert}, resulting in 13,404 gene expression measurements per sample. Finally, studies from the train set with compounds or chemicals known to induce liver injury were selected (n=74) to extract the 1,000 genes with the largest log2 fold change, used for our analysis. The log2 fold change gene expression values were not further normalized before processing by the deep learning system. In total, we obtained 6,597 transcriptomic samples used for training. 

\noindent\textbf{Histology data overview:} A summary of the liver data (TG-GATEs), Breast carcinoma (BRCA), and Lung carcinoma (NSCLC) is presented in Table~\ref{table:tggates_data_split}, Table~\ref{table:brca_data_split} and Table~\ref{table:nsclc_data_split}.

\begin{table}[ht]
\centering
\caption{\textbf{TG-GATEs data split overview.} Normal refers to benign WSIs without lesions. Positive refers to WSIs with lesions as reported by toxicologic pathologists.}
\begin{tabular}{lccc}
\toprule
 & {Samples} & {Normal} & {Positive} \\
 \midrule
iBOT-Tox pre-training & 47,227  & -- & -- \\
$\ours$ pre-training & 6,597  & 5,204 & 1,393 \\
Few-shot train & 2,783 & 2,322 & 461 \\
Independent test & 4,584 & 3,858 & 726 \\
\bottomrule
\end{tabular}
\label{table:tggates_data_split}
\end{table}

\begin{table}[ht]
\centering
\caption{\textbf{BRCA data split overview.} All (S+E) pre-training slides were included for few-shot training.}
\begin{tabular}{lccc}
\toprule
 & {Samples} & {IDC} & {ILC} \\
 \midrule
$\ours$ pre-training & 1,041  & 831 & 210 \\
Few-shot train & 1,041 & 831 & 210 \\
Independent test & 1,265 & 982 & 283 \\
\bottomrule
\end{tabular}
\label{table:brca_data_split}
\end{table}

\begin{table}[ht]
\centering
\caption{\textbf{NSLCL data split overview.} All (S+E) pre-training slides were included for few-shot training.}
\begin{tabular}{lccc}
\toprule
 & {Samples} & {LUAD} & {LUSC} \\
 \midrule
$\ours$ pre-training & 1,033  & 528 & 505 \\
Few-shot train & 1,033 & 528 & 505 \\
Independent test & 1,946 & 1,621 & 325 \\
\bottomrule
\end{tabular}
\label{table:nsclc_data_split}
\end{table}

\begin{table*}
  \centering
  \caption{
  \textbf{iBOT-Tox pre-training hyperparameters.} 8 $\times$ 80GB NVIDIA A100 GPUs were used for training. Batch size refers to the total batch size across GPUs.}
  \begin{tabular}{l|l}
    \toprule
    Hyperparameter & Value \\
    \midrule
    Layers & 12 \\
    Heads & 12 \\
    Patch size & 16 \\
    Head activation & GELU \\
    Embedding dimension & 768 \\
    Drop path rate & 0.1 \\
    \midrule
    Global crop scale & 0.32, 1.0 \\
    Global crop number & 2 \\
    Local crop scale & 0.05, 0.32 \\
    Local crop number & 10 \\
    Partial prediction shape & Block \\
    Partial prediction ratio & 0.0, 0.3 \\
    Partial prediction variance & 0, 0.2 \\
    Gradient clipping & 0.3 \\
    Normalize last layer & \checkmark \\
    Shared head & \checkmark \\
    \midrule
    AdamW $\beta$ & (0.9, 0.999) \\
    Batch size & 1024 \\
    Freeze last layer epochs & 3 \\
    Warmup epochs & 5 \\
    Warmup teacher temperature epochs & 30 \\
    Max epochs & 80 \\
    Learning rate schedule & Cosine \\
    Learning rate (start) & 0 \\
    Learning rate (post warmup) & 5e-4 \\
    Learning rate (final) & 2e-6 \\
    Teacher temperature (start) & 0.04 \\
    Teacher temperature (final) & 0.07 \\
    Teacher momentum (start) & 0.996 \\
    Teacher momentum (final) & 1.000 \\
    Weight decay (start) & 0.04 \\
    Weight decay (end) & 0.4 \\
    Automatic mixed precision & fp16 \\
    \bottomrule
  \end{tabular}
  \label{tab:hparams_ibot}
\end{table*}

\section{Results}

\noindent\textbf{Lesion-wise TG-GATEs few-shot performance:} To better understand $\ours$ few-shot performance on TG-GATEs for lesion classification in rat liver, we provide per-lesion classification performance, namely, on cellular infiltration, fatty change, (hepatocellular) hypertrophy, increased mitosis, (hepatocellular) necrosis, and proliferation of bile duct and oval cells. These lesions can take various sizes, \emph{e.g.,} necrosis can be focal (located in a small region) or diffuse (scattered all over the tissue). Lesions can also have different morphologies, such as hepatocellular hypertrophy that can be accompanied by eosinophilic or basophilic degeneration. As presented in Table~\ref{table:lesion-wise}, large lesions such as fatty change and hypertrophy are easier to detect than smaller ones like cellular infiltration and necrosis. This may be due to the expression profiles not expressing focal lesions, for instance, because the amount of tissue that includes the lesion of interest is too small. 

\begin{table*}[t]
\centering
\caption{\textbf{Lesion-wise few-shot linear probing performance of $\ours$ in rat liver.} $\ours$ is tested on an independent test cohort comprising 4,584 slides, without any data leakage (slide- or study-level) from unimodal and multimodal pre-training. Average AUC and standard deviation are reported over five runs.}
\begin{tabular}{l|ccccc}
\toprule
Lesion & $k$=1$(\uparrow)$ & $k$=5$(\uparrow)$ & $k$=10$(\uparrow)$ & $k$=25$(\uparrow)$ & $k$=50\footnotemark$(\uparrow)$ \\
\midrule
 Cellular infiltration & 56.9 $\pm$ 14.5 & 60.3 $\pm$ 14.1 & 69.8 $\pm$ 2.3 & 71.5 $\pm$ 3.2 & 74.9 $\pm$ 3.7 \\
 Fatty change & 74.6 $\pm$ 23.3 & 74.3 $\pm$ 21.5 & 89.8 $\pm$ 2.6 & 92.7 $\pm$ 1.8 & 94.6 $\pm$ 0.5 \\
 Hypertrophy & 84.6 $\pm$ 7.7 & 86.3 $\pm$ 10.4 & 90.0 $\pm$ 2.5 & 92.1 $\pm$ 1.5 & 91.3 $\pm$ 1.8 \\
 Increased mitosis & 75.5 $\pm$ 7.2 & 89.9 $\pm$ 2.9 & 89.7 $\pm$ 1.5 & 89.7 $\pm$ 1.1 & 89.7 $\pm$ 0.4 \\
 Necrosis & 56.4 $\pm$ 15.8 & 75.6 $\pm$ 5.9 & 74.9 $\pm$ 6.3 & 79.8 $\pm$ 2.0 & 78.1 $\pm$ 2.8 \\
 Proliferation & 84.4 $\pm$ 5.0 & 93.9 $\pm$ 2.6 & 94.0 $\pm$ 1.3 & 91.8 $\pm$ 2.3 & 92.8 $\pm$ 2.7 \\
\midrule
 Mean & 72.1 $\pm$ 11.6 & 80.1 $\pm$ 11.3 & 84.7 $\pm$ 9.0 & 86.3 $\pm$ 7.9 & 86.9 $\pm$ 7.6 \\
\bottomrule
\end{tabular}
\label{table:lesion-wise}
\end{table*}
\footnotetext{50 or maximal available labeled samples per class}

\begin{figure*}[t]
   \centering
   \includegraphics[width=\linewidth]{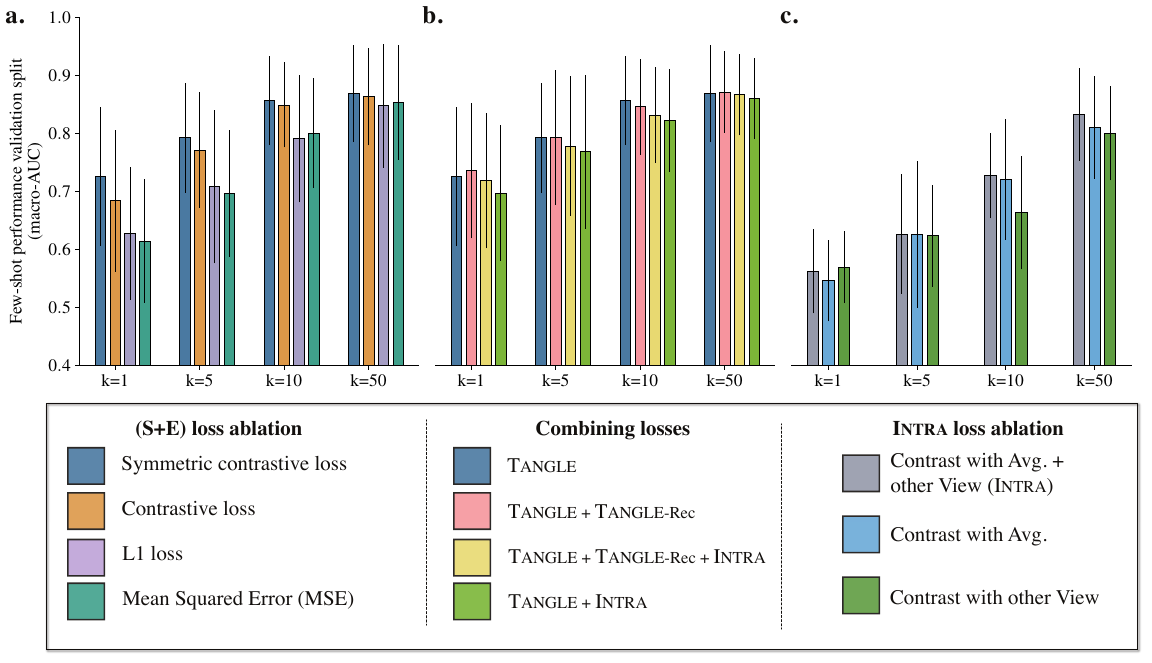}
   \caption{\textbf{Ablation study on TG-GATES.}
   \textbf{a.} Ablation of the (S+E) loss of $\ours$. We compare a symmetric contrastive loss with its non-symmetric counterpart, an L1 loss, and a Mean Squared Error loss. 
   \textbf{b.} Combining $\ours$ loss with $\ours$-Rec and $\textsc{Intra}$.
   \textbf{c.} $\textsc{Intra}$ loss ablation using the average patch embedding, a random other view based on a different patch set, or a combination of both. 
   }
   \label{fig:ablation_loss}
\end{figure*}

\noindent\textbf{Loss ablation:} We conduct three types of ablations on TG-GATEs: (1) ablation of the $\ours$ loss, (2) ablation of the $\textsc{Intra}$ loss, and (3) experiments where we combine $\ours$ and $\textsc{Intra}$ (see Figure~\ref{fig:ablation_loss}).

First, we compare the symmetric contrastive objective of $\ours$ with a one-sided objective (image $\rightarrow$ expression). Adding a symmetric loss leads to a consistent performance boost. We also tested with a mean-squared error (L2) and an L1 objective, both leading to a performance drop of 7.0\% and 6.7\% AUC, respectively. In addition, we compare the gain of using both a local-global and local-local contrastive alignment in $\textsc{Intra}$. Both objectives bring complementary information and lead to a performance loss when only one is employed. Finally, we combine $\ours$ objective with an $\textsc{Intra}$ objective based on contrasting the average token (Contrast w. Avg.) and a random view (Contrast w. Random View). Combining both leads to a performance drop of -2.0\% AUC. 

\begin{figure*}[t]
   \centering
   \includegraphics[width=0.5\linewidth]{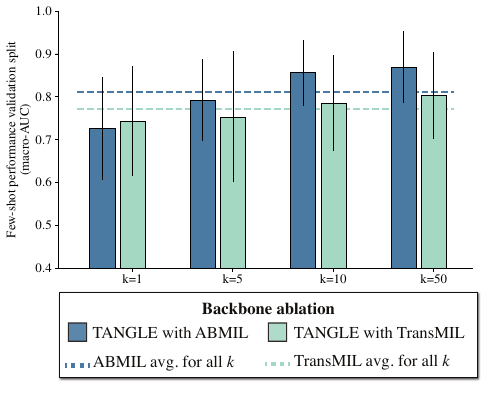}
   \caption{\textbf{Model ablation on TG-GATEs.} $\ours$ training when replacing the ABMIL backbone by TransMIL. 
   }
   \label{fig:ablation_model}
\end{figure*}

\noindent\textbf{Model ablation:} $\ours$ uses an attention-based MIL (ABMIL) as backbone. We compare the performance of $\ours$ when replacing it with TransMIL~\cite{shao2021transmil} (see Figure~\ref{fig:ablation_model}). This modification leads to a performance drop of 3.92\% AUC. We hypothesize that (1) the tasks we focus on (TG-GATEs lesion classification and TCGA lung and breast subtyping) are predominantly morphological, thereby reducing the utility of context modeling, (2) ABMIL training can use larger batch sizes due to reduced memory requirements; and (3) our ABMIL implementation uses ``modern tricks'' such as a deeper pre-attention network and LayerNorm (see implementation).  

\begin{figure*}[t]
   \centering
   \includegraphics[width=\linewidth]{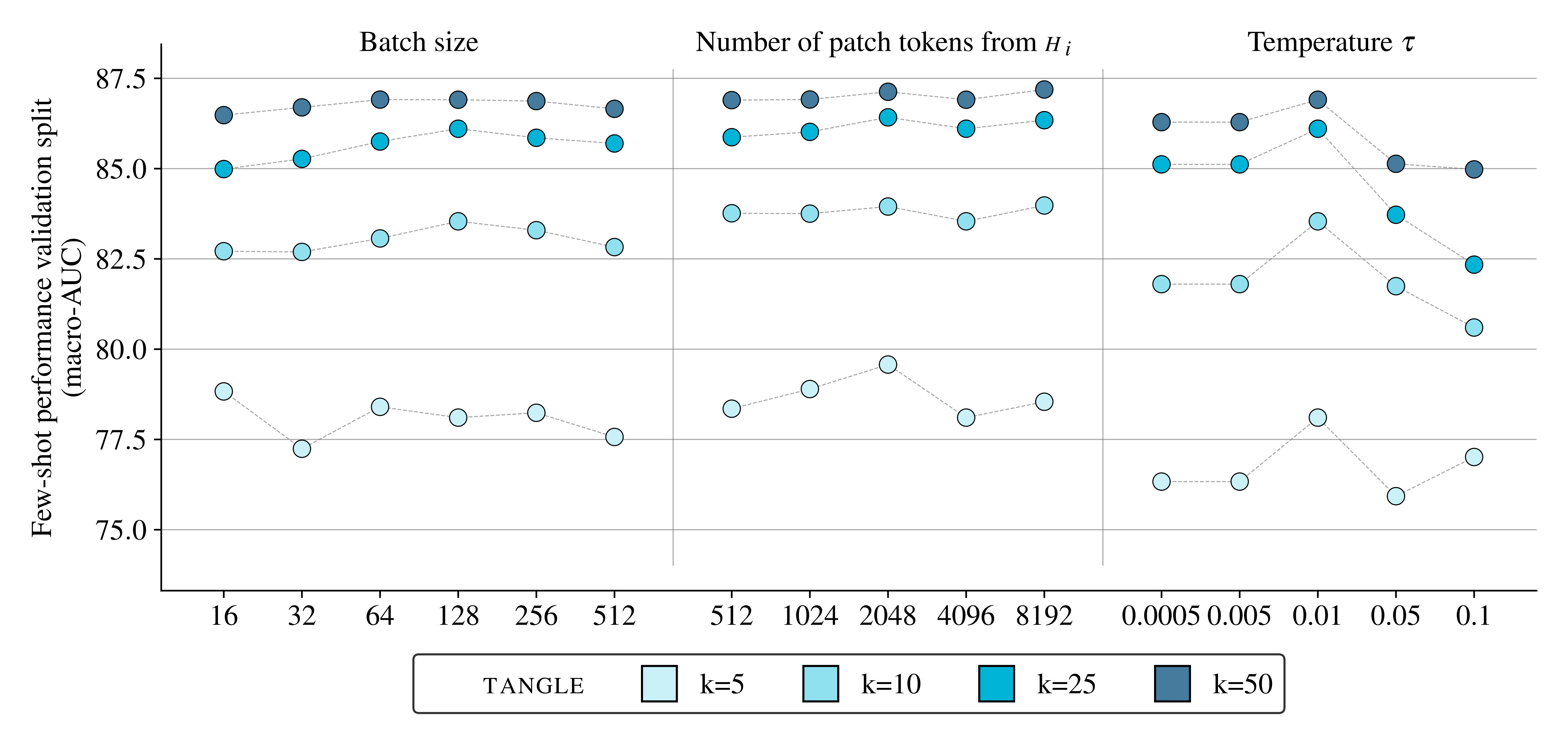}
   \caption{\textbf{Hyper-parameter search on TG-GATES.} We assess the influence of the batch size, number of patches sampled per slide, and the Softmax temperature.}
   \label{fig:hyper}
\end{figure*}

\noindent\textbf{Hyper-parameter search:} Figure~\ref{fig:hyper} presents a series of experiments with different hyper-parameters known to influence contrastive pre-training, namely, the batch size, the Softmax temperature, and the number of patches sampled per slide. Batches larger than 64 seem to perform equally well. Softmax temperatures that are too high lead to a severe performance drop. Finally, the number of tokens (or patches) sampled per slide has relatively little influence on the downstream few-shot performance. 

\section{Interpretability}

\begin{figure*}[t]
    \centering
   \includegraphics[width=0.85\linewidth]{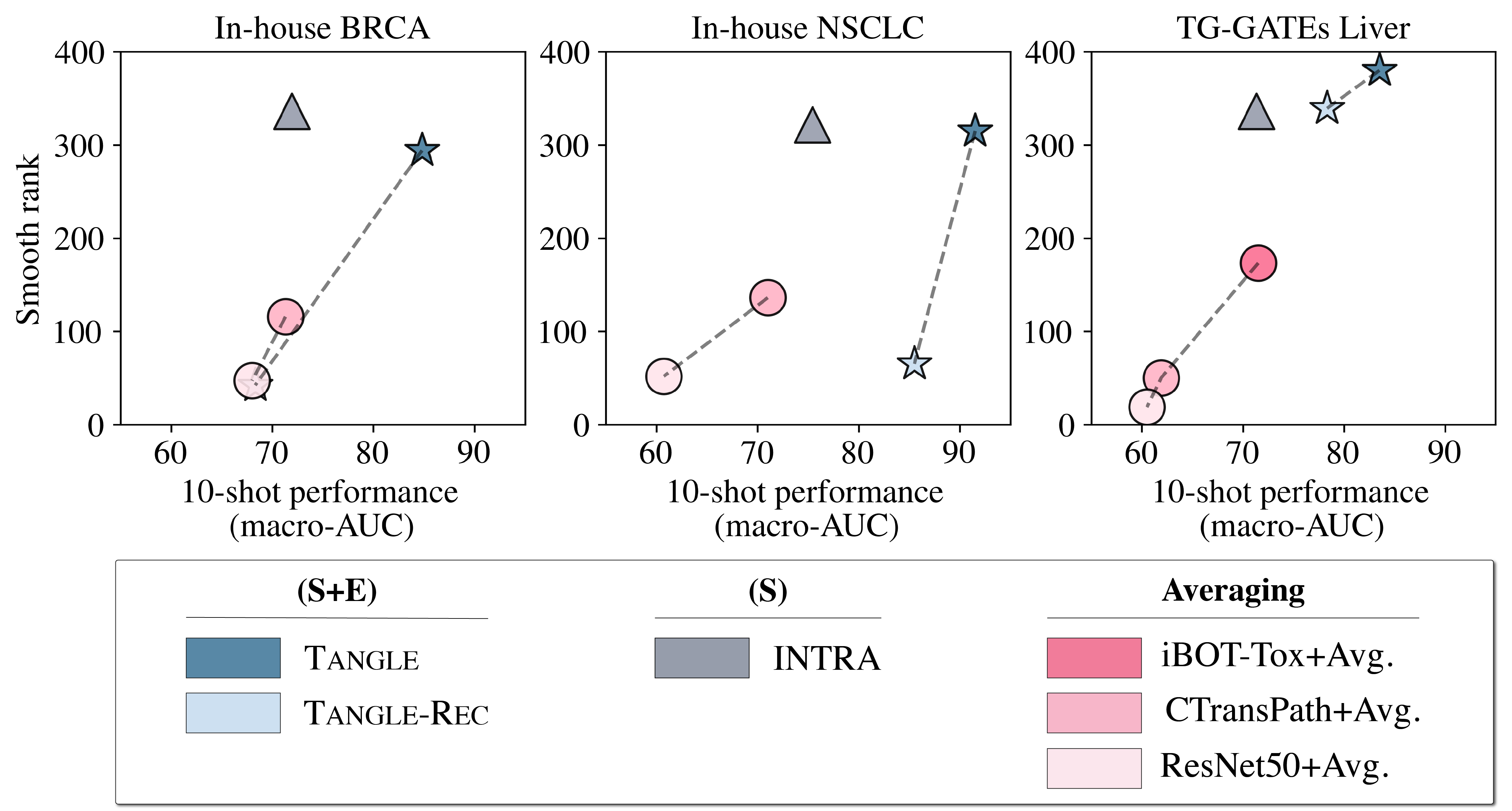}
   \caption{\textbf{Few shot performance \emph{vs.} smooth rank.} $\ours$ linear probing performance ($k$=10) and baselines, plotted against the smooth rank of the slide embedding matrix of the independent test cohorts. Test cohorts tested on BRCA subtyping (human breast, n=1,265 WSIs), NSCLC subtyping (human lung, n=1,946 WSIs), and TG-GATEs lesion classification (rat liver, n=4,584 WSIs). For each family of methods, we observe a strong positive correlation between performance and rank. 
   }
   \label{fig:smooth_rank}
\end{figure*}

\noindent\textbf{Rank analysis:} Following~\cite{garrido2022rankme}, we use the rank as a fast and cheap measure of the quality of the underlying latent space learned during SSL pre-training. Here, we compute the rank as the entropy of the $d$ (assuming $d<n$) L1-normalized singular values of the slide embedding matrix $H \in \real^{n \times d}$. Specifically, we have:
\begin{align}
    \text{RankMe}(H) &= \exp(-\sum_{k=1}^d p_k\log(p_k))\;, \\
    p_k &= \frac{\sigma_k(H)}{|\sigma(H)|_1} + \epsilon
\end{align}
where $\sigma_k$ denotes the $k-$th singular of $H$ (sorted from large to low), and $\epsilon$ is small constant set to $1e-7$ for numerical stability. Figure~\ref{fig:smooth_rank} presents the smooth rank of the slide embeddings obtained with different methods on the three independent test cohorts. 

\begin{figure*}[ht]
\centering
   \includegraphics[width=0.8\linewidth]{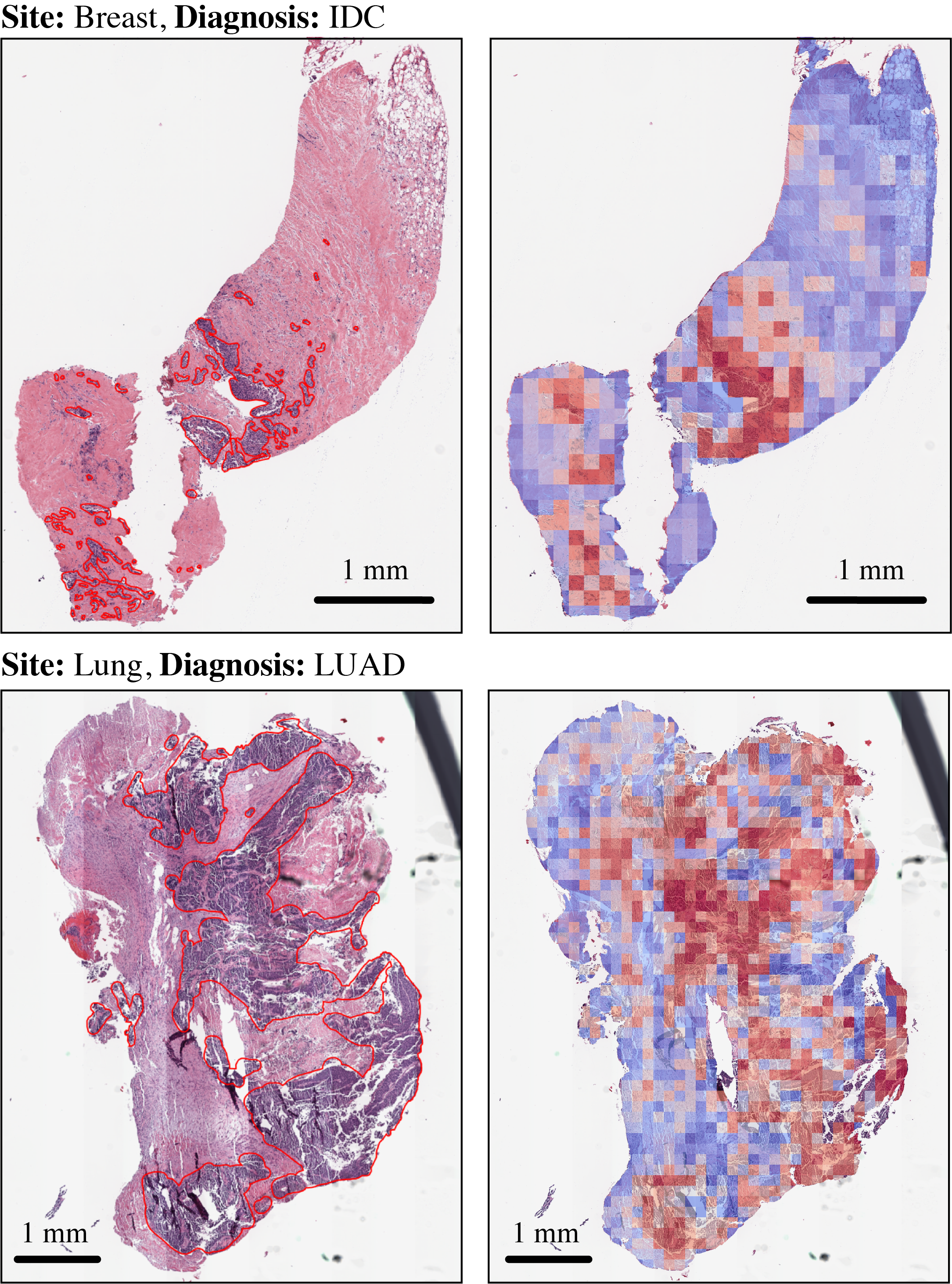}
   \caption{\textbf{$\ours$ attention heatmaps of a lung and breast slide.} Attention weights of the (frozen) ABMIL slide encoder pre-trained with $\ours$ overlaid on randomly chosen samples for our in-house cohorts. The network focuses mostly on tumor regions (marked in red) in both the breast and lung samples. This is a remarkable property of (S+E) pre-training as the network was not explicitly trained for tumor-related tasks, such as subtyping or grading.}
   \label{suppfig:mgb_heatmaps}
\end{figure*}

\noindent\textbf{Attention heatmaps:} We also present attention heatmaps of $\ours$ when pre-trained on breast (Figure~\ref{suppfig:mgb_heatmaps}, top) and lung (Figure~\ref{suppfig:mgb_heatmaps}, bottom). Interestingly, the attention is assigned to regions that overlap with tumor, a property that naturally emerges from multimodal pre-training without explicit training. 

\end{document}